\documentclass[sn-mathphys-num]{sn-jnl}
\usepackage{graphicx}
\usepackage{multirow}
\usepackage{amsmath,amssymb,amsfonts}
\usepackage{amsthm}
\usepackage{mathrsfs}
\usepackage[title]{appendix}
\usepackage{xcolor}
\usepackage{textcomp}
\usepackage{manyfoot}
\usepackage{booktabs}
\usepackage{algorithm}
\usepackage{algorithmicx}
\usepackage{algpseudocode}
\usepackage{listings}
\usepackage{graphicx}
\usepackage{subcaption}
\usepackage{fix-cm}
\usepackage{mathrsfs}
\usepackage{caption}
\usepackage{ulem}

\begin{document}

\title[Article Title]{Enhancing knowledge retention for continual learning with domain-specific adapters and features gating\footnote{This manuscript has been submitted to Applied Intelligence (Springer) and has been under review since November 26, 2024.}}

\author[1]{\fnm{Mohamed Abbas Hedjazi} \sur{}}\email{abbas.hedjazi@dasia.ai}
\equalcont{These authors contributed equally to this work.}

\author[1]{\fnm{Oussama Hadjerci} \sur{}}\email{oussama.hadjerci@dasia.ai}
\equalcont{These authors contributed equally to this work.}

\author[2]{\fnm{Adel Hafiane} \sur{}}\email{adel.hafiane@insa-cvl.fr}

\affil[1]{\orgname{DASIA}, \orgaddress{\street{25 Quai du Prés. Paul Doumer}, \city{Courbevoie}, \postcode{92400},  \country{France}}}

\affil[2]{\orgdiv{INSA CVL}, \orgname{PRISME}, \orgaddress{\street{88 Bd Lahitolle}, \city{Bourges}, \postcode{18000}, \country{France}}}

\abstract{Continual learning empowers models to learn from a continuous stream of data while preserving previously acquired knowledge, effectively addressing the challenge of catastrophic forgetting and preserving original abilities. In this study, we propose a new approach that integrates adapters within the self-attention mechanisms of Vision Transformers to enhance knowledge retention when sequentially adding datasets from different domains. Unlike previous methods, which continue learning with only one dataset, our approach introduces domain-specific output heads and features gating, allowing the model to maintain high accuracy on previously learned tasks while seamlessly incorporating only the essential information of more than one domain.
The proposed method is compared to the prominent parameter-efficient-fine-tuning methods in the current state-of-the-art. The result provide further evidence that our method can effectively alleviate the limitation of previous works. Furthermore, we conduct a comparative analysis using three datasets: CIFAR-100, Flowers102, and DTD, each representing a distinct domain, to investigate the impact of task order on model performance. Our findings underscore the critical role of dataset sequencing in shaping learning outcomes, demonstrating that strategic ordering can significantly improve the model's ability to adapt to evolving data distributions over time while preserving the integrity of previously learned knowledge.
}

\maketitle

\section{Introduction}
\label{section:introduction}

Continual learning \cite{wang2024comprehensive} is an emerging paradigm in artificial intelligence that focuses on training models to learn new information while maintaining the accuracy of previous knowledge. By allowing models to learn incrementally and update themselves over time, continual learning opens up possibilities for applications in dynamic environments, such as robotics \cite{lesort2020continual}, autonomous systems \cite{verwimp2023clad}, healthcare \cite{amrollahi2022leveraging}, and personalized recommendations \cite{cai2022reloop}. Such systems need to evolve and respond to new data in real time, improving their performance without requiring frequent retraining from scratch.

Traditional machine learning models, typically trained on a single static dataset, often encounter significant challenges in real-world applications where data evolves over time. One of the foremost issues is catastrophic forgetting \cite{de2021continual, kemker2018measuring}, a phenomenon where new learning interferes with and overwrites previous knowledge, severely limiting the model's overall performance and usability. Various strategies have been proposed to mitigate catastrophic forgetting in continual learning settings, including regularization methods \cite{chen2021overcoming, el2019preempting}, rehearsal-based techniques \cite{verwimp2021rehearsal, evron2022catastrophic}, and architectural expansion \cite{zhang2020one}. Regularization and rehearsal-based approaches, while helpful, face challenges with scalability and computational efficiency, particularly when dealing with large, complex datasets over time \cite{sokar2021spacenet}. Similarly, architectural expansion methods, though effective at reducing forgetting, often result in increased model complexity and resource demands, limiting their practicality in real-world applications \cite{wang2023comprehensive}. 
\begin{figure}[h]
    \centering
    \includegraphics[width=0.95\linewidth]{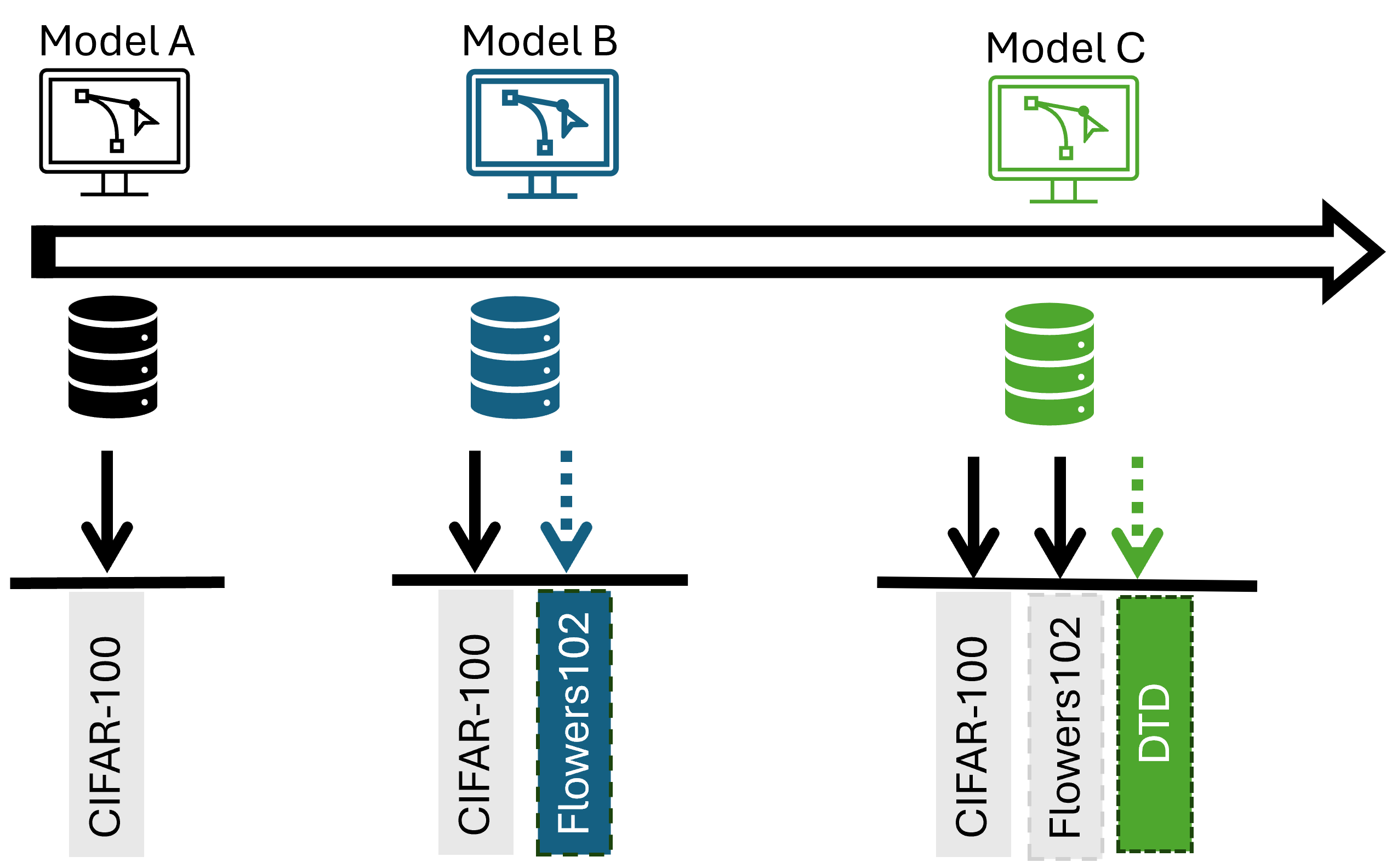} 
    \caption{Continual domain learning.}
    \label{fig:idea}
\end{figure}
Recently, \cite{AkbarianBafghi2024ParameterEF} tackled the issue of catastrophic forgetting in Vision Transformers (ViTs) \cite{dosovitskiy2020image}, but its approach is constrained by the use of only a single fine-tuning dataset, limiting the broader applicability and generalizability of its results. 
To overcome this limitation, we propose a new approach to continual learning using a dynamic architecture. Our method integrates low-rank adaptation (LoRA) \cite{LoRA2021} into the self-attention mechanisms of ViTs, allowing domain-specific parameters to be introduced while preserving the original weights from previously learned tasks. Thus facilitating knowledge retention across distinct domains. 
To enhance this retention, we also incorporate domain-specific output heads and features gating, enabling the model to maintain high accuracy on prior tasks while effectively learning new ones.
Furthermore, we expand the evaluation by including multiple datasets (Figure~\ref{fig:idea}) and examining how task order impacts model performance, providing insights into the balance between stability and adaptability. In particular, we assess the effect of dataset sequence permutations on model performance, evaluating the efficacy of our proposed architecture using three distinct datasets: CIFAR-100 \cite{Cifar1002009}, Flowers102 \cite{Flowers1022014}, and DTD \cite{DTD2010}. Through extensive experimentation, we explore various permutations and their impact on accuracy, aiming to identify optimal configurations for continual learning. Our results show that the sequence in which datasets are presented significantly influences learning outcomes, underscoring the importance of strategic task ordering in the training process. The key contributions of this paper are as follows:

\begin{itemize}
    \item We propose a new approach for continual learning, incorporating a dynamic architecture with domain-specific output heads and features gating that leverages LoRA layers within ViTs to enhance knowledge retention.
    \item We adapt parameter-efficient fine-tuning techniques for continual learning, emphasizing their structural modifications within this framework.
    \item We conduct an in-depth analysis of how the sequence of dataset training impacts model performance, providing actionable insights for optimizing continual learning strategies.
    \item We provide insights into optimal configuration, highlighting the role of model architecture and comparing our method with parameter-efficient fine-tuning techniques adapted for continual learning.
\end{itemize}

\section{Related Works}
\label{section:related_works}

\subsection{Large Vision Models}
Large vision models have become pivotal in the field of computer vision, significantly advancing performance across various tasks such as image classification \cite{araujo2020automated} and object detection \cite{li2022exploring}. 
These architectures revolutionized the way visual data is processed, leading to substantial improvements in accuracy and efficiency. The introduction of ViTs \cite{dosovitskiy2020image} marked a significant shift in the landscape of computer vision by adapting the transformer architecture \cite{vaswani2017attention} to visual data. ViTs have demonstrated the ability to outperform traditional CNNs like ResNet \cite{he2016deep}, especially when trained on large datasets. This shift highlights the versatility of transformer-based models in capturing complex relationships within images. To enhance model performance in downstream tasks, I-JEPA \cite{assran2023self} and DINO \cite{caron2021emerging} utilize self-supervised learning to capture robust and meaningful features without labeled data. By improving feature representations in ViTs, these approaches expand the applicability and effectiveness of vision models across diverse contexts. Our method proposes a novel approach that builds upon DINO, further enhancing its ability to generate high-quality feature representations for continual learning. 

\subsection{Continual learning}
Continual learning aims to develop models that can adapt over time while retaining previously acquired knowledge, making it essential for applications in dynamic environments where conditions and requirements frequently change. One of the primary challenges faced in this field is catastrophic forgetting, a phenomenon where new learning interferes with and overwrites previously learned information. To address this issue, researchers have proposed various strategies. Regularization methods \cite{chen2021overcoming, el2019preempting} are among the most prominent approaches, adding constraints to the learning process to preserve important weights associated with earlier tasks. While effective in many situations, regularization methods can struggle under particularly challenging settings, where the complexities of new tasks may lead to significant knowledge loss. Additionally, these methods may introduce extra computational overhead and can require careful tuning of hyperparameters to be effective. Rehearsal-based methods \cite{verwimp2021rehearsal, evron2022catastrophic} involve retaining a subset of previous data to rehearse during training, helping the model maintain its performance on earlier tasks. However, these methods can be limited by memory constraints, as they may require storing a large amount of historical data, which can be impractical in resource-constrained environments. Architectural approaches \cite{zhang2020one} also offer solutions by creating models with dedicated components that allow for better separation of domain-specific knowledge. Nonetheless, these architectures can increase model complexity and may require significant adjustments to accommodate new tasks effectively \cite{sokar2021spacenet}. In our approach, we use architectural methods but address these limitations by incorporating adapter modules, which introduce fewer parameters compared to traditional architectural changes.

\subsection{Parameter Efficient Fine-Tuning}
Parameter-efficient fine-tuning (PEFT) is essential in continual and transfer learning, as it enables models to adapt to new tasks with minimal retraining. In this study, we compare our approach with three prominent methods: Prefix Tuning \cite{li2021prefix}, Block Expansion \cite{Wu2024LLaMAPP} and LoRA \cite{LoRA2021}. 

\subsubsection{Prefix tuning}
In prefix tuning \cite{li2021prefix, Jia2022VisualPT}, instead of fine-tuning the entire model, a set of task-specific embeddings are learned and prepended to the input sequence for each task. Let \( X \in \mathbb{R}^{n \times d} \) represent the original input embeddings, where \( n \) is the sequence length and \( d \) is the embedding dimension. A sequence of prefix embeddings \( P \in \mathbb{R}^{m \times d} \) is learned and prepended to the input \( X\), to obtain \( X'\in \mathbb{R}^{(m+n) \times d} \), where \( m \) is the length of the prefix. Only the parameters of \( P \) are optimized during training, while the model's core weights remain frozen. This approach allows the model to adapt to various tasks with minimal additional parameters, as only \( P \) needs to be learned for each task.

\subsubsection{Block Expansion}

Block Expansion \cite{Wu2024LLaMAPP} is a technique that increases the capacity of pre-trained ViTs without altering their initial outputs. Given a set of transformer blocks \( \{\phi_0, \phi_1, \dots, \phi_N\} \), block expansion adds an identity block \( \phi_{\text{id}} \) such that \(\phi_{\text{id}}(x) = x\). This ensures that the output remains the same after the identity block is added. To expand the model from \( N \) to \( N' \) blocks, the original blocks are grouped into sets containing \( M \) blocks. Within each set, an identity copy of the topmost block is added, leading to the new set:
\[
\phi_0, \phi_1, \dots, \phi_{M-1}, \phi_{id}^{1}, \phi_{M}, \dots, \phi_N\, \phi_{id}^{k}
\]

The new identity blocks are initialized with zero-initialized linear layers to enable identity mapping. These added blocks are then fine-tuned with the new data, while the original blocks remain frozen.

\subsubsection{LoRA}
LoRA \cite{LoRA2021} is a low-rank adaptation method applied into the self-attention blocks of the transformer architecture. In LoRA, the weight matrix \( W \in \mathbb{R}^{d \times k} \) of a pre-trained model is decomposed as:
\[
W + \Delta W = W + \frac{\alpha}{r} A B
\]
where \( A \in \mathbb{R}^{d \times r} \) and \( B \in \mathbb{R}^{r \times k} \) are low-rank matrices with rank \( r \), and \( \alpha \) is a scaling factor. Here, \( \Delta W = \frac{\alpha}{r} A B \) represents the learned adaptation, allowing LoRA to adjust task-specific parameters without modifying the original weights \( W \). The low-rank structure \( A \) and \( B \) significantly reduces the number of parameters required for adaptation.

\subsection{Features gating}

Previous works on conditional computation in deep learning have explored dynamic filtering and gating strategies to improve model performance and efficiency. For instance, the GaterNet \cite{Chen2018YouLT} introduces input-dependent dynamic filter selection in convolutional neural networks, leading to improved generalization and interpretability. Similarly, the autors in \cite{Bejnordi2019BatchShapedCG}, propose fine-grained gating for individual convolutional maps, reducing computational cost while enhancing accuracy. Another approach called Gated convolution was introduce in \cite{Yu2018FreeFormII}. This work is based on learning soft masks for dynamic feature selection in tasks like image inpainting. We adapt this technique to continual learning, applying gating mechanisms to selectively retain mandatory features learned from previous domains. Specifically, we extend this idea to LoRA layers, allowing dynamic feature selection for new tasks, thereby ignoring unnecessarily information from previous tasks.

\section{Methodology}
\label{section:methodology}

Our approach leverages LoRA and features gating to achieve domain-specific adaptation without compromising foundational knowledge, using domain-specific output heads that adapt to unique dataset characteristics. This configuration enhances computational efficiency by reducing the need to update extensive parameters, thereby supporting efficient adaptation to new tasks.

\begin{figure}[h]
    \centering
    \includegraphics[width=0.40\linewidth]{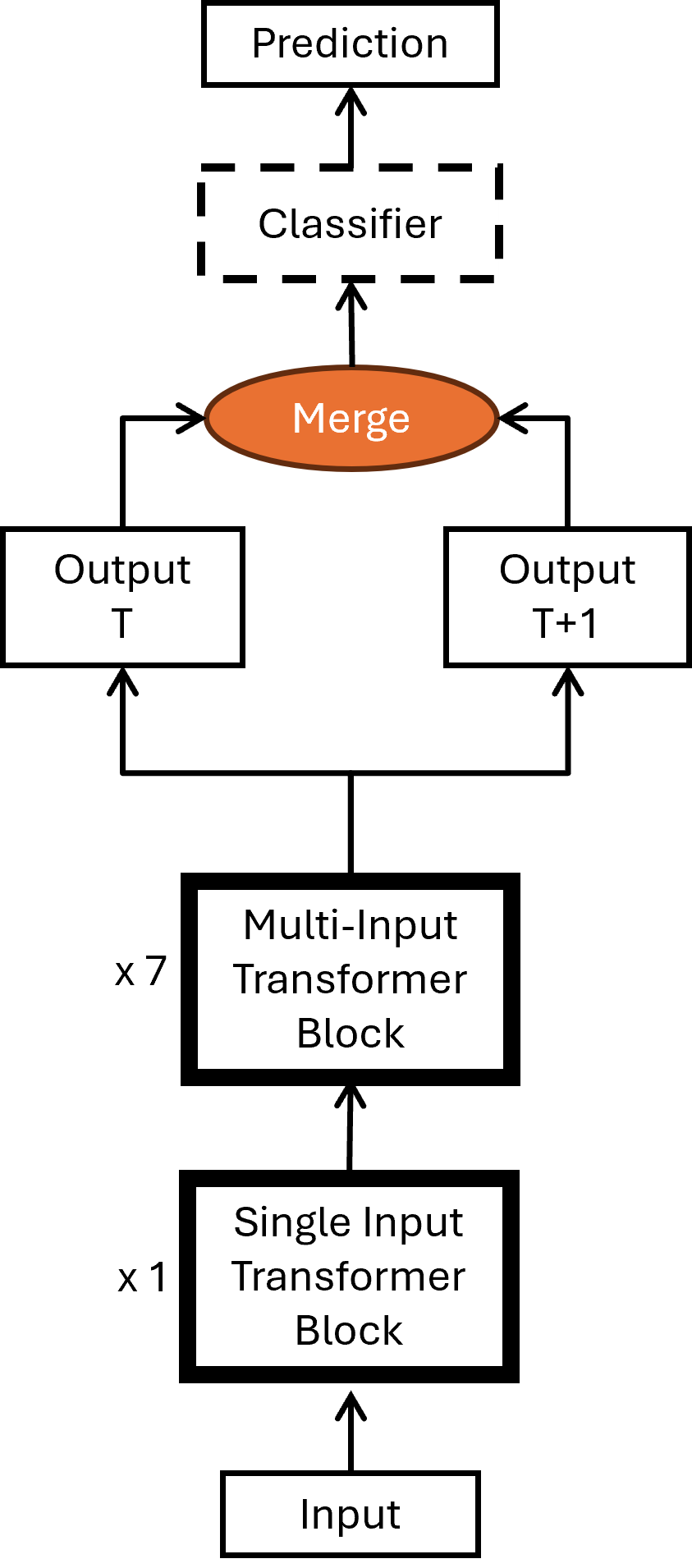} 
    \caption{Our proposed architecture with multiple outputs.}
    \label{fig:architecture_general}
\end{figure}
\subsection{Model Architecture}

Our proposed architecture builds upon the transformer framework and introduces modifications aimed at enhancing continual learning. As illustrated in Figure~\ref{fig:architecture_general}, the input is fed to eight transformer blocks to give multiple outputs where each one corresponds to both previous (output T) and current (output T+1) distributions. Before classification, these outputs are aggregated (merged) by averaging across domains distributions. 
\begin{figure}[h]
    \centering
    \begin{subfigure}[b]{0.36\linewidth}
        \centering
        \includegraphics[width=0.69\textwidth]{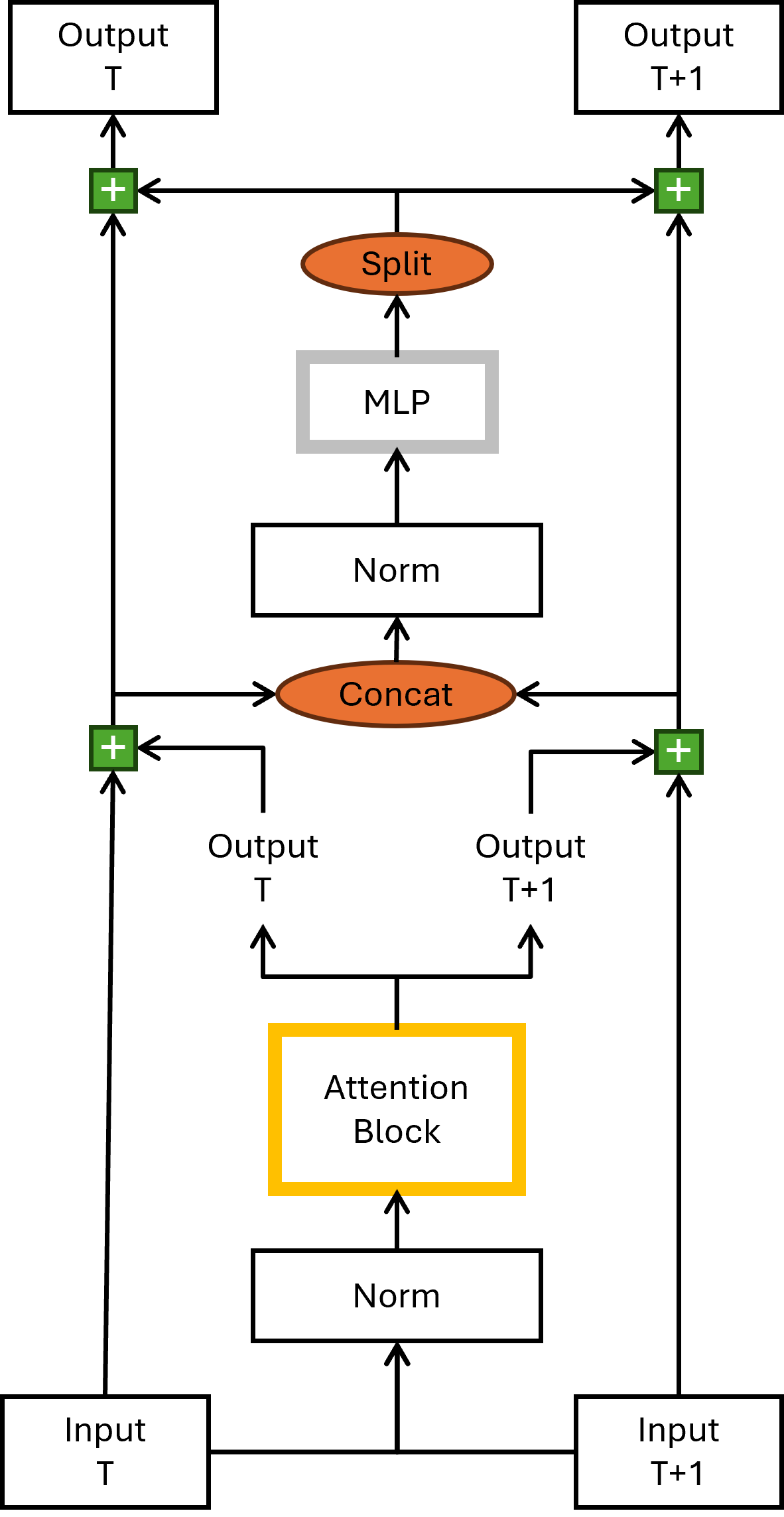}
        \caption{Single input block}
        \label{fig:architecture_a}
    \end{subfigure}
    \begin{subfigure}[b]{0.36\linewidth}
        \centering
        \includegraphics[width=0.69\textwidth]{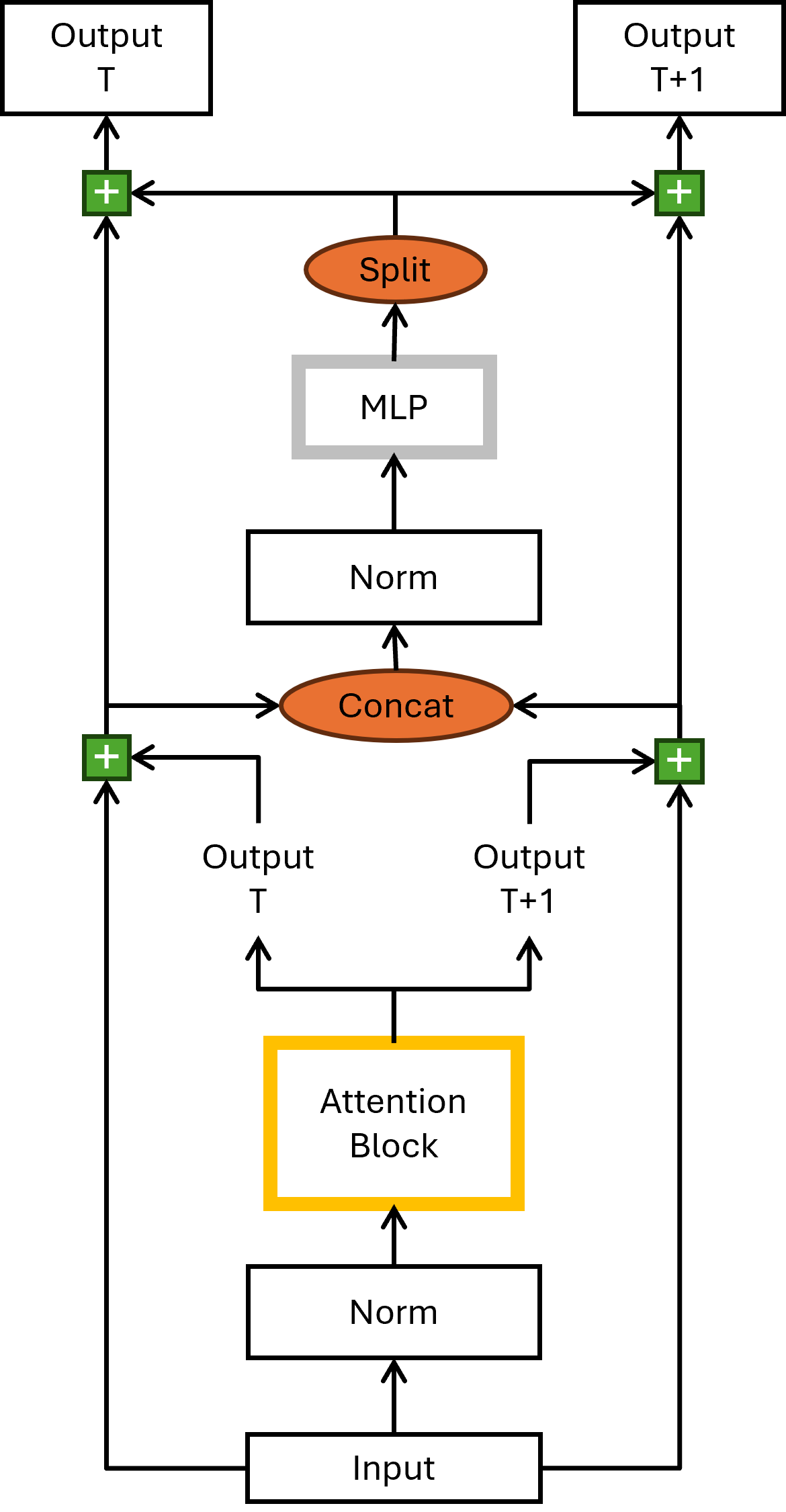}
        \caption{Multiple inputs block}
        \label{fig:architecture_b}
    \end{subfigure}
    \begin{subfigure}[b]{0.65\linewidth}
        \centering
        \includegraphics[width=0.9\textwidth]{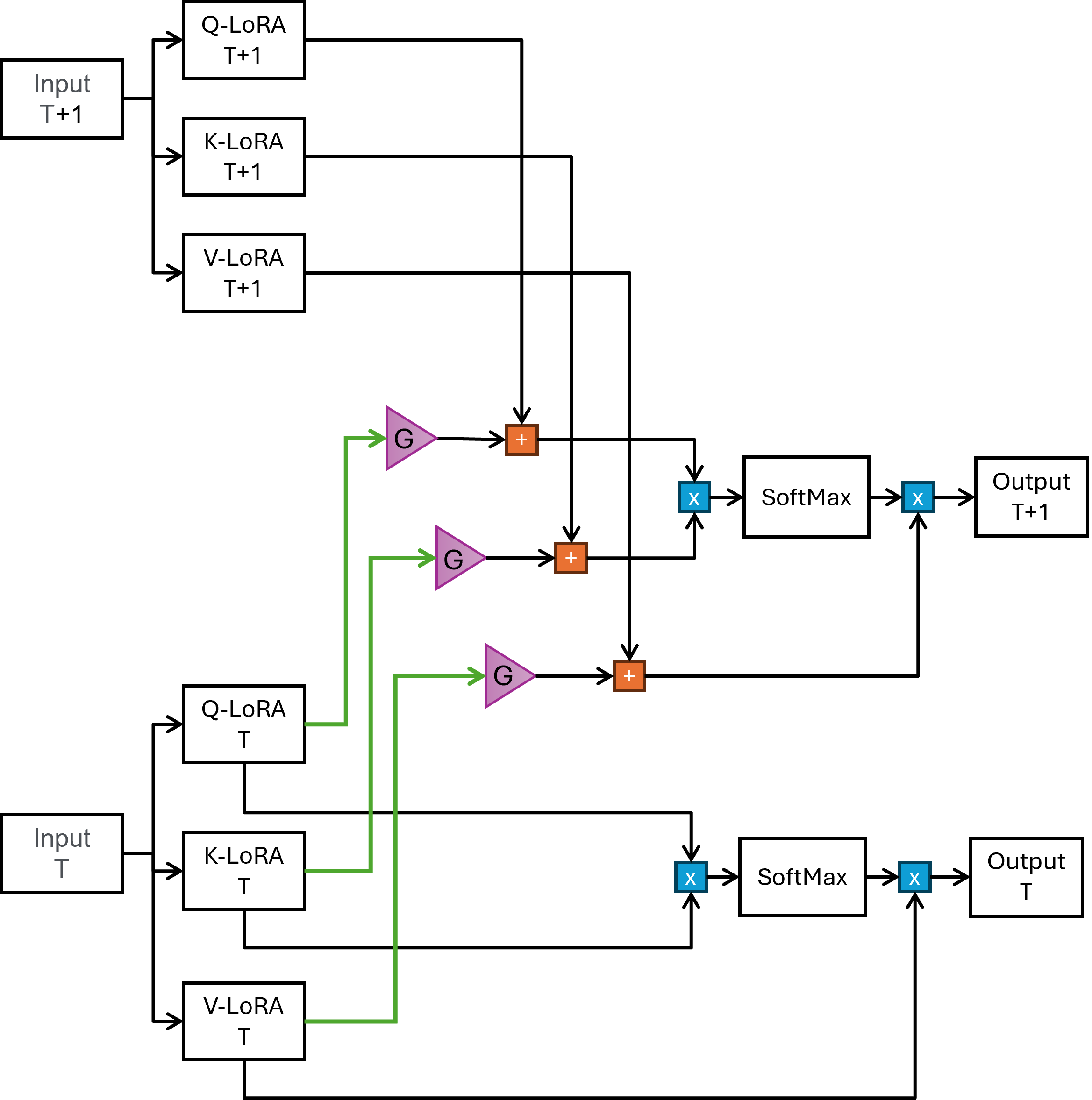}
        \caption{Attention block}
        \label{fig:architecture_c}
    \end{subfigure}
    \caption{Detailed illustration of different architectural components.}
    \label{fig:architecture}
\end{figure}
Figure~\ref{fig:architecture} details a transformer block that resembles a traditional one \cite{vaswani2017attention}, with added capabilities for handling both single (Figure~\ref{fig:architecture_a}) and multiple inputs (Figure~\ref{fig:architecture_b}). The architecture's first block takes a single input and produces multiple outputs, each corresponding to a different domain distribution. Then, in later blocks (Figure~\ref{fig:architecture_b}), multiple inputs are normalized and given to the attention block (Figure~\ref{fig:architecture_c}) to yield multiple outputs. These are concatenated, and normalized before passing through a multi-layer perceptron (MLP). The final MLP output is partitioned by the number of domain distributions, allowing the model to effectively capture features from past and current tasks. 

In Figure~\ref{fig:architecture_c}, we present our modified attention block. Initially, inputs (T and T+1) undergo projection through both the previous domain distribution: query (\(Q^{T}_{LoRA}\)), key (\(K^{T}_{LoRA}\)), and value (\(V^{T}_{LoRA}\)) projectors, and the new domain distribution: \(Q^{T+1}_{LoRA}\), \(K^{T+1}_{LoRA}\), and \(V^{T+1}_{LoRA}\) projectors. Post-projection, \(Q_{LoRA}\), \(K_{LoRA}\), \(V_{LoRA}\)for T and T+1 are summed after the gating mechanism (G) (see Section~\ref{section:methodology_gating}). For each distribution, Q and K are multiplied, followed by the application of the SoftMax function to normalize the attention scores, and then multiplied by V, the attention mechanism is defined as:

\[
\text{Attention}(Q, K, V) = \text{softmax} \left( \frac{QK^T}{\sqrt{d_k}} \right) V
\]

Where \( d_k \): is the dimensionality of the head vector.

During training, we freeze the Q, K, and V components of previous domain distributions to retain past knowledge, thus preventing catastrophic forgetting and enhancing learning efficiency across diverse datasets by using features gating technique. This approach enables faster adaptation by building on previous insights while maintaining learned patterns.
\subsection{Features gating}
\label{section:methodology_gating}
\begin{figure}[h]
    \centering
    \includegraphics[width=0.25\linewidth]{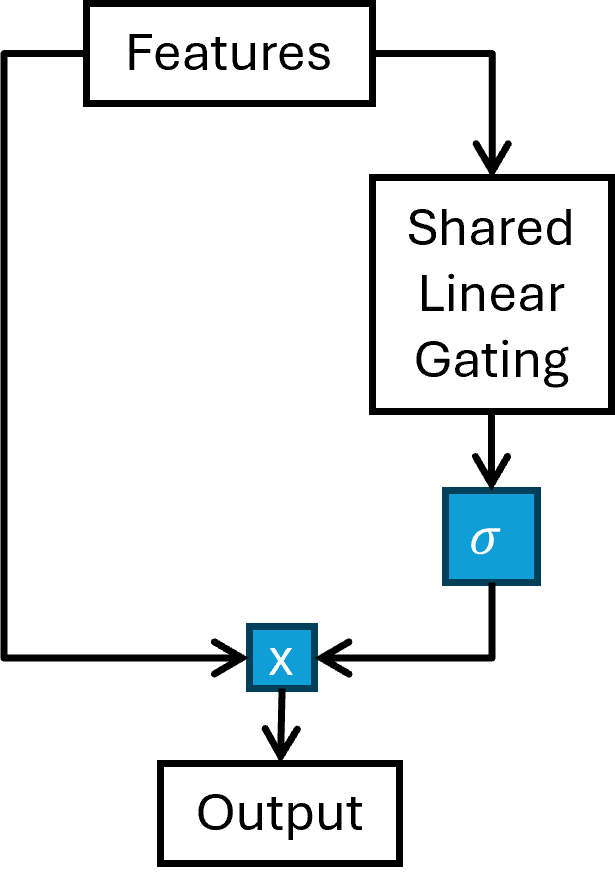} 
    \caption{Gating mechanism (G).}
    \label{fig:architecture_gate}
\end{figure}

In our approach we add a gating mechanism (Figure~\ref{fig:architecture_gate}) to ensure that corrupted information existing in the previous domain does not propagate into the next domain distribution.
First, the gating values for the previous domain is computed by applying the shared linear layers \(W_{g}\) to \(Q^{T}_{LoRA}, K^{T}_{LoRA}, V^{T}_{LoRA}\), followed by a sigmoid activation function \(\sigma\):
    \[
    Gating_{Q} = \sigma(W_{g} \cdot Q^{T}_{LoRA}) 
    \]
    \[
    Gating_{K} = \sigma(W_{g} \cdot K^{T}_{LoRA})
    \]
    \[
    Gating_{V} = \sigma(W_{g} \cdot V^{T}_{LoRA})
    \]
The final output is the multiplication of \(Q^{T}_{LoRA}, K^{T}_{LoRA}, V^{T}_{LoRA}\) by the gating values, then summed to the next task domain \(Q^{T+1}_{LoRA}, K^{T+1}_{LoRA}, V^{T+1}_{LoRA}\).
    \[
    Output_{Q} = Gating_{Q} \cdot Q^{T}_{LoRA} + Q^{T+1}_{LoRA}
    \]
    \[
    Output_{K} = Gating_{K} \cdot K^{T}_{LoRA} + K^{T+1}_{LoRA}
    \]
    \[
    Output_{V} = Gating_{V} \cdot V^{T}_{LoRA} + V^{T+1}_{LoRA}
    \]


\subsection{Domain-specific output heads}
As illustrated in Figure~\ref{fig:architecture_b}, our approach assigns a dedicated output head to each dataset, effectively preventing catastrophic forgetting and supporting continual learning. By using separate LoRA layers, each domain has its specific output head. Hence, we avoid parameter overwriting and retain knowledge from prior tasks. This structure enables the model to leverage shared features across tasks while tailoring outputs to each dataset's specific requirements. The dedicated heads provide clear task boundaries, allowing the model to integrate new knowledge without compromising performance on previous datasets, enhancing adaptability and resilience in continual learning.

\section{Experimental settings}
\label{section:experiments}
In this section, we first define the datasets used, then outline the implementation details of our approach and finally, we explain our integration of prefix tuning and block expansion for continual learning.

\subsection{Datasets}
\begin{figure}[h]
    \centering
    \begin{subfigure}[b]{0.36\linewidth}
        \centering
        \includegraphics[width=0.8\textwidth]{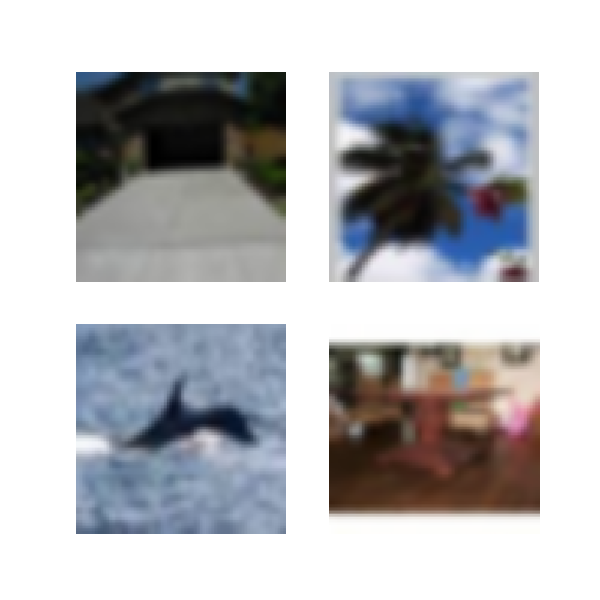}
        \caption{CIFAR-100}
        \label{fig:dataset_cifar100}
    \end{subfigure}
    \begin{subfigure}[b]{0.36\linewidth}
        \centering
        \includegraphics[width=0.8\textwidth]{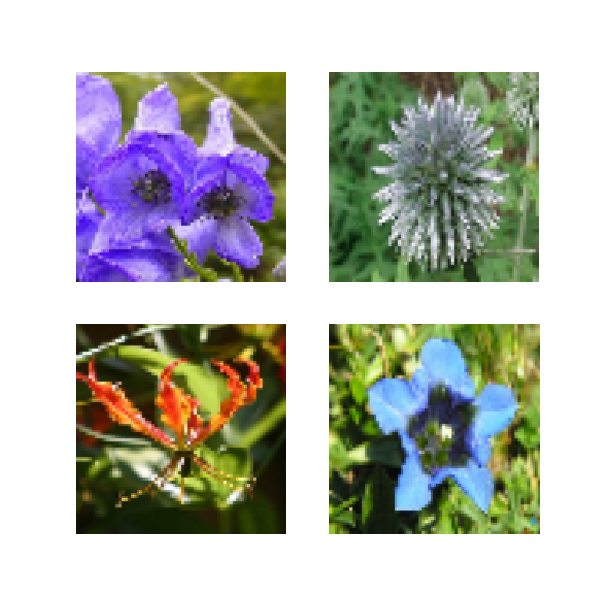}
        \caption{Flowers102}
        \label{fig:dataset_flowers102}
    \end{subfigure}
    \begin{subfigure}[b]{0.36\linewidth}
        \includegraphics[width=0.8\textwidth]{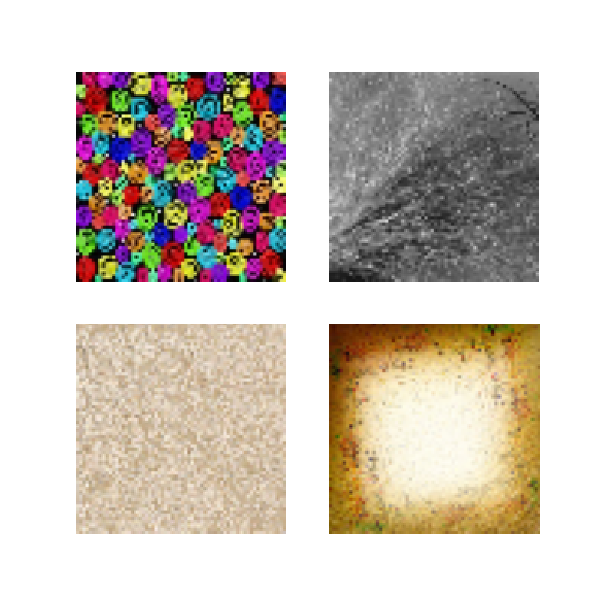}
        \caption{DTD}
        \label{fig:dataset_ctd}
    \end{subfigure}
    \caption{Illustration of visual differences between data domains.}
    \label{fig:dataset_samples}
\end{figure}
We conduct experiments on three distinct datasets, each selected due to its unique domain characteristics. CIFAR-100 \cite{Cifar1002009} contains a wide range of objects across various categories, Flowers102 \cite{Flowers1022014} focuses on objects within a single domain, and DTD \cite{DTD2010} consists of textures rather than objects. These diverse datasets allow us to assess the impact of continual learning across different domains and to understand the effect of knowledge forgetting when transitioning between tasks. As illustrated in Figure~\ref{fig:dataset_samples}, we provide visual samples that highlight these differences.

\begin{itemize}
    \item \textbf{CIFAR-100:} This dataset is a subset of 60000 tiny RGB images with a resolution of $32 \times 32$ pixels, containing 100 classes. It presents a diverse range of objects, making it suitable for evaluating the model's ability to generalize across different categories.
    
    \item \textbf{Flowers102:} This dataset contains a total of 8189 RGB images of size $224 \times 224$ pixels with 102 categories of flowers, characterized by variations in scale, pose, and lighting conditions. It serves as a challenging benchmark for assessing the model's performance in recognizing fine-grained visual differences.
    
    \item \textbf{DTD:} This texture dataset comprises 5640 RGB images of size $224 \times 224$ pixels, categorized into 47 distinct texture patterns based on human perception. It provides a unique opportunity to evaluate the model's understanding of texture recognition and its ability to differentiate between various surface patterns.
\end{itemize}

The preprocessing steps for all datasets include resizing images to a uniform size, normalization to standardize pixel values, and application of data augmentation techniques such as random cropping, rotation, and flipping. These techniques aim to enhance the model's robustness by exposing it to a broader range of visual variations, thereby improving its generalization capabilities across different tasks.

\subsection{Implementation details}

In this section, we outline the key implementation details of our proposed approach, including the initial setup of the model, the training process, and the performance evaluation metric. 

\begin{itemize}
    \item \textbf{Initial setup:} The model is initialized with DINO-pretrained weights specifically ViT-S and ViT-B. We configure the LoRA layers with specified ranks of 8 and 16. Also, We use the sigmoid activation function in the features gating. 
    
    For optimization, we employ the Stochastic Gradient Descent (SGD) optimizer \cite{sebastian2016overview}, with an initial learning rate of 0.01 and a weight decay of zero to prevent overfitting. To dynamically adjust the learning rate during training, we apply a cosine learning rate scheduler that gradually decays the learning rate to 0.001, promoting stable convergence.

    \item \textbf{Training process:} The model is trained on each dataset sequence for a total of 25 epochs on an NVIDIA RTX A6000 GPU. Throughout the training process, we monitor the validation accuracy and select the epoch with the highest one for subsequent evaluation. This approach ensures that performance assessment is based on the model's peak accuracy during training, maximizing its effectiveness.

    \item \textbf{Training sequences:} We evaluate various dataset sequences to understand how training order affects continual learning. By testing different combinations, we aim to identify the most effective sequence for model performance, focusing on knowledge retention and adaptability to new information. This analysis will provide insights into how domains distribution influences learning dynamics and helps minimize catastrophic forgetting.
    \item \textbf{Performance evaluation:} For fairness evaluation of continual learning techniques, we utilize the k-nearest neighbors (KNN) method to evaluate the fine-tuned model’s performance on feature representations. Three metrics are used: (1) accuracy, to indicate the performance model at the end of training. (2) positive backward transfer, to highlight the impact of learning new task to reinforce the knowledge of prior task and (3) forward transfer, to measure how knowledge gained from learning previous tasks helps to learn new tasks.

\end{itemize}

\subsection{Adaptation of PEFTs for continual learning}
\label{sec:experimental_adaptation}

Through these adaptations, we conduct a thorough comparison of our proposed architecture against prefix tuning \cite{li2021prefix}, block expansion \cite{Wu2024LLaMAPP} and LoRA \cite{LoRA2021}, focusing on their structural modifications in the continual learning context.

\subsubsection{Prefix Tuning}
For each dataset, we define a sequence of prefix embeddings \( P_{i} \in \mathbb{R}^{m \times d} \), corresponding to dataset \( i \) (with \( i \in \{1, 2, 3\} \) for CIFAR-100, DTD, and Flowers102). The resulting modified input embeddings for each dataset are:

\begin{itemize}
\item CIFAR-100: \( X'_{1} = \begin{bmatrix} P_{1} \\ X \end{bmatrix} \in \mathbb{R}^{(m+n) \times d} \)

\item DTD:
\(
X'_{2} = \begin{bmatrix} P_{2} \\ X \end{bmatrix} \in \mathbb{R}^{(m+n) \times d}
\)

\item Flowers102:
\(
X'_{3} = \begin{bmatrix} P_{3} \\ X \end{bmatrix} \in \mathbb{R}^{(m+n) \times d}
\)
\end{itemize}

During training, only the parameters of  \( P_{i} \) are optimized, while the core weights of the model remain frozen.

\subsubsection{Block Expansion} 

In the context of continual learning, we extend the block expansion technique to sequentially incorporate knowledge from multiple datasets, namely CIFAR-100, Flowers102, and DTD. 

For each new dataset, we add a single block expansion after the existing transformer blocks \( \{\phi_0, \phi_1, \dots, \phi_N\} \). For each subsequent dataset, an identity block \( \phi_{\text{id}} \) is added after the existing blocks to expand the model’s capacity as follows:

\[
\{\phi_0, \phi_1, \dots, \phi_N, \phi_{\text{id}}^{\text{CIFAR-100}}, \phi_{\text{id}}^{\text{Flowers102}}, \phi_{\text{id}}^{\text{DTD}}\}
\]

where \( \phi_{\text{id}}^{\text{CIFAR-100}} \), \( \phi_{\text{id}}^{\text{Flowers102}} \), and \( \phi_{\text{id}}^{\text{DTD}} \) are the identity blocks specific to each dataset.

\subsubsection{LoRA} 
We represent the LoRA equations with task-specific matrices \( A_i \) and \( B_i \) for CIFAR-100, Flowers102, and DTD. Given a pre-trained weight matrix \( W_0 \in \mathbb{R}^{d \times d} \), LoRA introduces low-rank updates for each task \( i \) with a scaling factor \( \alpha \). The effective weight for each task \( i \) is defined as:
\[
W_i = W_0 + \Delta W_i = W_0 + \alpha_i A_i B_i
\]
Hence, for each dataset, we have:
\[
CIFAR100: \Delta W_1 = \alpha_{1} A_1 B_1
\]
\[
Flowers102: \Delta W_2 = \alpha_{2} A_2 B_2
\]
\[
DTD: \Delta W_3 = \alpha_{3} A_3 B_3
\]
The final weight after applying all updates from the sequence of datasets is:
\[
W_{\text{final}} = W_0 + \Delta W_1 + \Delta W_2 + \Delta W_3 = W_0 + \alpha_1 A_1 B_1 + \alpha_2 A_2 B_2 + \alpha_3 A_3 B_3
\]

\section{Results and discussion}
\label{section:results}

In this section, we conduct an ablation study of different components in our approach, then we compare it against PEFT methods. Throughout these experiences, we provide insights into the dynamics of continual learning. 

\subsection{Ablation study}
In this section, we determine the best parameter $k$ for KNN and the optimal rank selection for LoRA, while also choosing the most suitable architecture. Meanwhile, we study the impact of domain distribution order, and we highlight the effect of features gating.
\subsubsection{Impact of number of Nearest Neighbors}

\begin{figure}[h]
    \centering
    \includegraphics[width=1.0\linewidth]{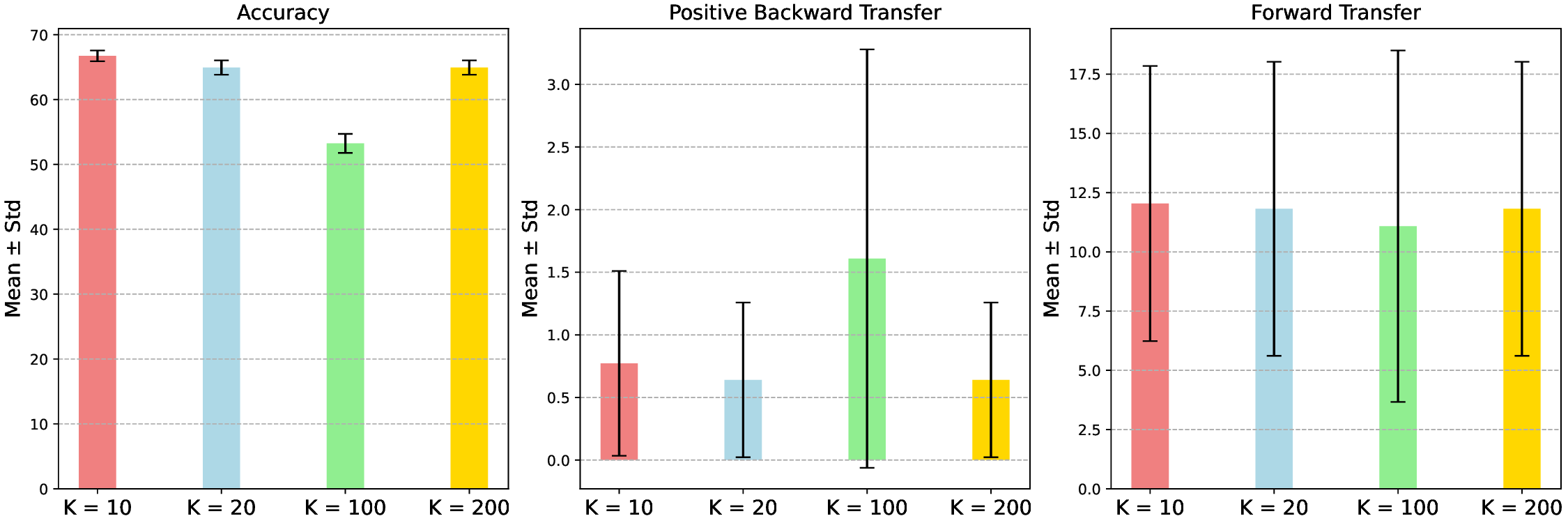}
    \caption{Impact of parameter $K$ on CL metrics using mean and std across all sequences.}
    \label{fig:ablation_all_metrics_mean_std}
\end{figure}

In the experiment shown in Figure~\ref{fig:ablation_all_metrics_mean_std}, we compare the accuracy obtained from four different values of $k \in (10, 20, 100, 200)$ and demonstrate that the $k=10$ configuration yields stable scores across all continual learning metrics (accuracy, positive backward transfer, and forward transfer) on all datasets sequences. While $K=100$ provides good mean positive backward transfer results, it exhibits a high variation, indicating that the performance is sensitive to sequence ordering.

\subsubsection{Optimal rank selection for LoRA}
\begin{figure}[h]
    \centering
    \includegraphics[width=1.0\linewidth]{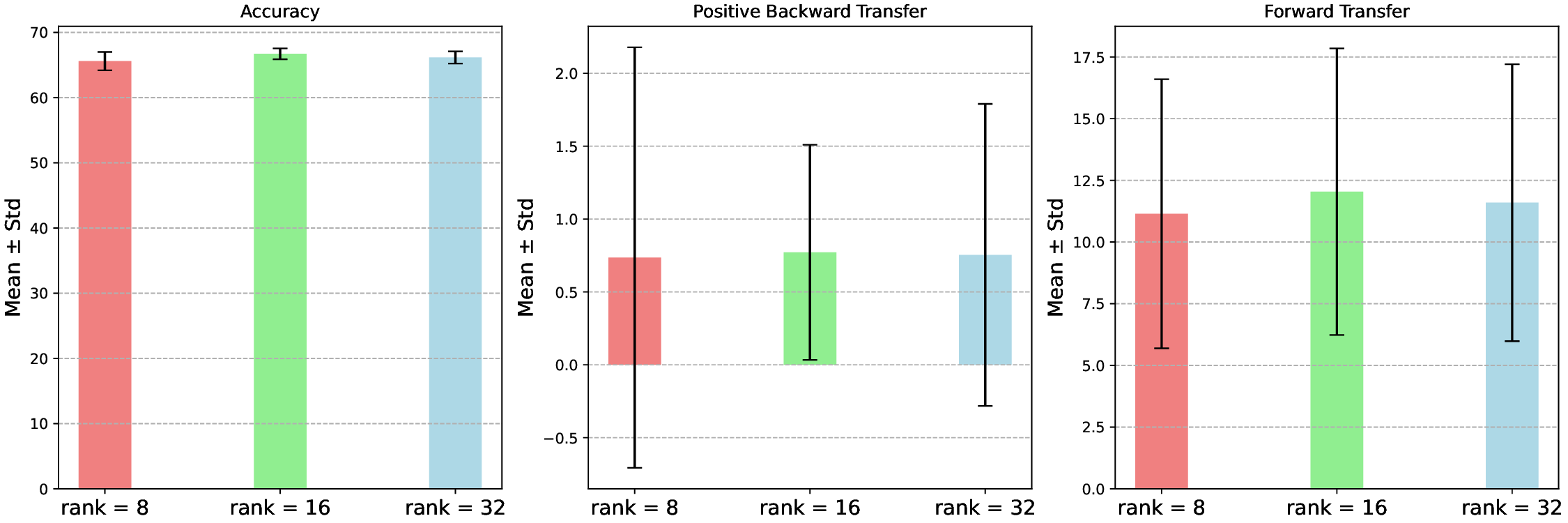}
    \caption{Rank comparison using CL metrics.}
    \label{fig:ablation_rank}
\end{figure}

The variation in ranks for LoRA layers significantly impacts overall model performance, highlighting that optimal rank selection is crucial for the model's adaptability. The results in Figure~\ref{fig:ablation_rank}, show that using a rank of 16 consistently yields better performance compared to a rank of 8. Although rank of 32 produces competitive results than a rank of 16 but with a higher number of parameters. This finding suggests that 16 is the best configuration, because it offers a balance between parameter efficiency and performance improvement, allowing the model to better handle the complexities of continual learning. Moreover, we note that the variation for $rank=8$ is notably high in terms of positive backward transfer and forward transfer. This variability arises because the performance is highly sensitive to the sequence ordering, which can significantly influence the results. This variability highlights the challenge of optimizing the model's performance in dynamic learning environments, where domains order plays a critical role in the model's ability to generalize and retain knowledge across tasks.

\subsubsection{Comparison of ViT-S and ViT-B}
As demonstrated on previous experiments, $k=10$ and $rank=16$ are the optimized choices on continual learning performance. Also, we study the influence of the model size and show its implications.
As shown in Table~\ref{tab:vits_vitb}, we compare the performance of the ViT-S and ViT-B models using our proposed architecture. The ViT-B model consistently outperforms the ViT-S model across almost all dataset sequences, demonstrating superior accuracy. Despite its larger parameters size and increased inference time, ViT-B provides a clear advantage in performance. While ViT-S is more efficient in terms of inference time and parameter size, with 22 million parameters compared to ViT-B's 85 million, the performance gap in favor of ViT-B underscores its capability to handle more complex tasks effectively.
\begin{table}[h]
\centering
\caption{Comparison of ViT-S and ViT-B  on CIFAR-100, Flowers102, DTD.}
\begin{tabular}{@{}lccc@{}} 
\toprule
Sequence & ViT-S (\%) & ViT-B (\%) \\ \midrule
CIFAR-100 $\rightarrow$ Flowers102 $\rightarrow$ DTD & 72.72/64.6/57.07 & 75.18/67.08/58.51 \\ 
CIFAR-100 $\rightarrow$ DTD $\rightarrow$ Flowers102 & 72.87/63.62/57.45  & 75.09/66.56/57.90 \\ 
Flowers102 $\rightarrow$ CIFAR-100 $\rightarrow$ DTD & 75.55/60.01/57.13  & 76.73/64.03/56.44 \\ 
Flowers102 $\rightarrow$ DTD $\rightarrow$ CIFAR-100 & 75.97/58.86/51.49  & 77.12/66.97/58.72 \\ 
DTD $\rightarrow$ CIFAR-100 $\rightarrow$ Flowers102 & 74.66/63.1/56.28  & 75.36/69.30/58.94 \\ 
DTD $\rightarrow$ Flowers102 $\rightarrow$ CIFAR-100 & 76.34/55.52/50.64  & 75.74/64.55/56.76 \\ 
\toprule
\end{tabular}
\label{tab:vits_vitb}
\end{table}

\subsubsection{Impact of domain order on model performance}
The results in Table~\ref{tab:results} demonstrate that the training order of domains in continual learning has a significant impact on model performance. 
\begin{table}[h]
\centering
\caption{Results of our Approach on CIFAR-100, Flowers102 and DTD with different sequence order.}
\begin{tabular}{@{}llcc|c@{}}
\toprule
Sequence & CIFAR-100 (\%) & Flowers102 (\%) & DTD (\%) & Mean (\%) \\ \midrule
CIFAR-100 $\rightarrow$ Flowers102 $\rightarrow$ DTD & 75.18 & 67.08  & 58.51 & \textbf{66.26} \\
CIFAR-100 $\rightarrow$ DTD $\rightarrow$ Flowers102 & 75.09 & 66.56  & 57.90 & \textbf{66.18} \\\midrule
Flowers102 $\rightarrow$ CIFAR-100 $\rightarrow$ DTD & 76.73 & 64.03  & 56.44 & 65.73 \\
Flowers102 $\rightarrow$ DTD $\rightarrow$ CIFAR-100 & 77.12 & 66.97  & 58.72 & 67.60 \\\midrule
DTD $\rightarrow$ CIFAR-100 $\rightarrow$ Flowers102 & 75.36 & 69.30  & 58.94 & 67.20 \\
DTD $\rightarrow$ Flowers102 $\rightarrow$ CIFAR-100 & 75.74 & 64.55  & 56.76 & 65.68 \\ \midrule
\end{tabular}
\label{tab:results}
\end{table}
Initially, accuracy scores show that beginning with CIFAR-100 leads to a stable overall performance since it contains diverse classes. In contrast, starting with Flowers102 or DTD, respectively, yields a lot of variation ($1,87\%$ and $1,52\%$). This suggests that starting with a generic dataset helps the model develop a stronger foundation for fine-grained datasets (DTD and Flowers102), enabling it to better handle more complex tasks later in training. The findings highlight the advantage of progressively increasing task complexity, allowing the model to build on its knowledge as training progresses.

\subsubsection{The effect of features gating}

The results in Table~\ref{tab:fg} highlight the impact of incorporating features gating (FG) on the model’s continual learning performance across three datasets: CIFAR-100, Flowers102, and DTD. Overall, the inclusion of FG consistently enhances performance across all tested sequence orders, underscoring its efficacy in previous features selection. For most dataset sequences, FG achieves noticeable accuracy gains compared to the baseline without FG. For instance, in the CIFAR-100 $\rightarrow$ Flowers102 $\rightarrow$ DTD sequence, the FG-enabled model yields an accuracy improvement of approximately $0.3\%$ across all datasets. This trend is similar in other sequences, where FG provides comparable or better accuracy across different stages. These improvements suggest that FG enhances knowledge retention and feature selection, particularly when transitioning to new datasets.

\begin{table}[h] 
\centering
\caption{Comparison of our approach w/wo Features Gating (FG) on CIFAR100, Flowers102, and DTD.}
\begin{tabular}{@{}lccc@{}} 
\toprule
Sequence & w/o FG (\%) & w/ FG (\%) \\ \midrule
CIFAR-100 $\rightarrow$ Flowers102 $\rightarrow$ DTD & 75.18/67.08/58.51 & 75.50/67.50/58.80 \\ 
CIFAR-100 $\rightarrow$ DTD $\rightarrow$ Flowers102 & 75.09/66.56/57.90 & 75.40/66.90/58.20 \\ 
Flowers102 $\rightarrow$ CIFAR-100 $\rightarrow$ DTD & 76.73/64.03/56.44 & 77.00/64.30/56.70 \\ 
Flowers102 $\rightarrow$ DTD $\rightarrow$ CIFAR-100 & 77.12/66.97/58.72 & 77.40/67.20/59.00 \\ 
DTD $\rightarrow$ CIFAR-100 $\rightarrow$ Flowers102 & 75.36/69.30/58.94 & 75.60/69.50/59.10 \\ 
DTD $\rightarrow$ Flowers102 $\rightarrow$ CIFAR-100 & 75.74/64.55/56.76 & 76.00/64.80/57.00 \\ 
\toprule
\end{tabular}
\label{tab:fg}
\end{table}

\subsection{Comparative study of our approach against PEFTs}

As mentioned in Section~\ref{sec:experimental_adaptation}, we conduct experiments using different dataset sequences. For a fair comparison, we use the same parameters obtained previously, ViT-B as model, rank $16$ for LoRA and $k=10$ for KNN. 

\begin{figure}[h]
    \centering
    \includegraphics[width=1.0\linewidth]{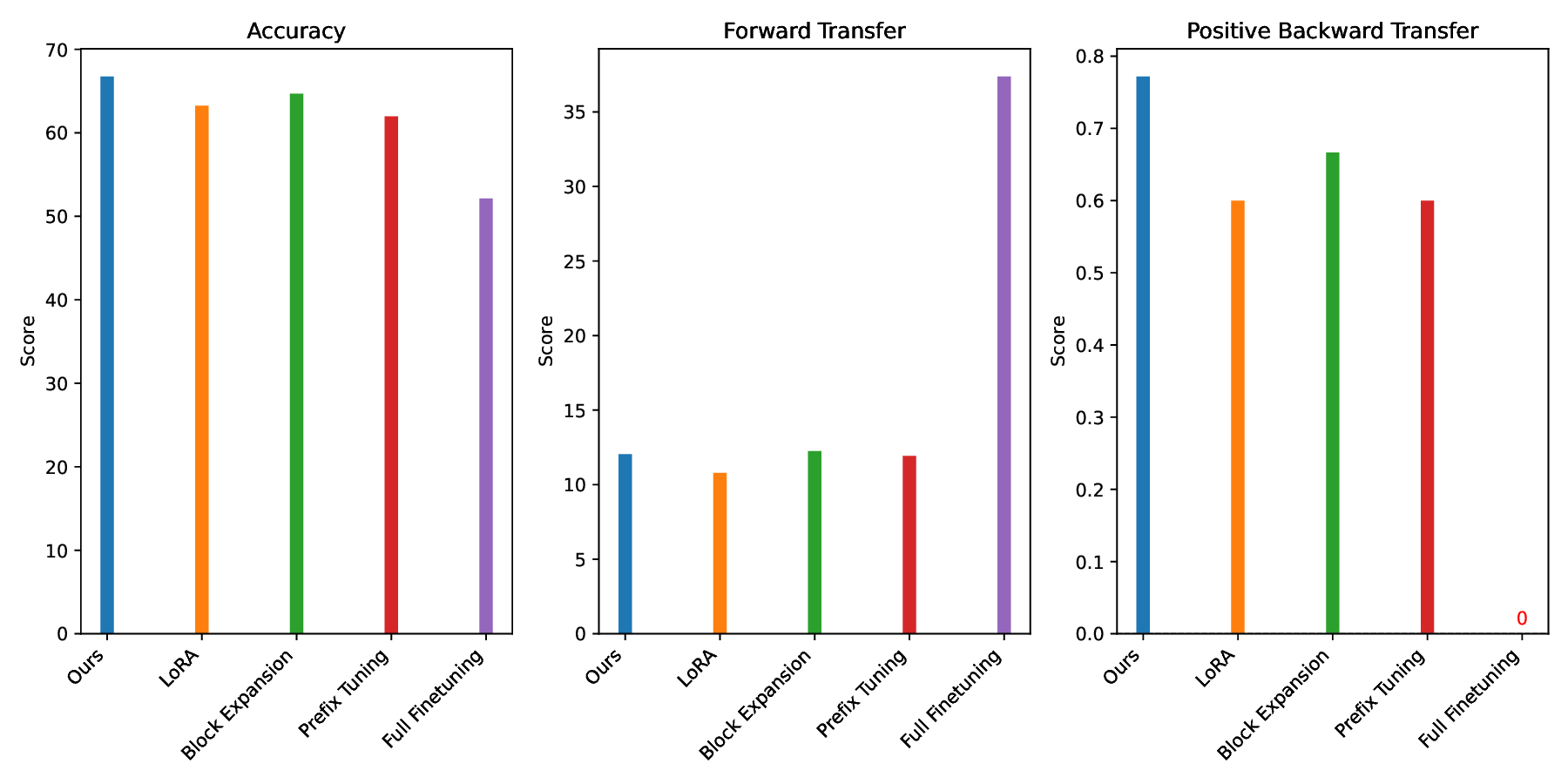}
    \caption{Comparison of PEFT state-of-the-art approaches using CL metrics.}
    \label{fig:sota_comparison}
\end{figure}

As shown in Figure~\ref{fig:sota_comparison}, our method outperformed Full Fine-tuning, Prefix Tuning \cite{li2021prefix}, Block Expansion \cite{Wu2024LLaMAPP} and LoRA \cite{LoRA2021} across all dataset sequences, achieving the highest accuracy, forward transfer and backward transfer. This strong performance highlights the ability of our method to effectively adapt pre-trained models to new tasks (forward transfer). While full finetuning generally suffers from catastrophic forgetting and is prone to degraded backward transfer when learning new tasks, it is an exception in terms of forward transfer. Full fine-tuning can, in some cases, provide excellent forward transfer due to its ability to fully adapt the model’s parameters to the new task. However, this benefit is often task-dependent, and it comes at the cost of performance on earlier tasks.

In contrast, our method preserves core model parameters while incorporating domain-specific modifications, ensuring superior forward transfer while maintaining knowledge retention (backward transfer). This makes it particularly well-suited for continual learning, where it is essential to maintain performance across multiple tasks.

In comparison, Block Expansion and LoRA offered competitive but slightly lower performance than our method. LoRA efficiently adapts models to new domains by leveraging low-rank decomposition, introducing fewer additional parameters than Block Expansion. However, it faced limitations in retaining accuracy across domains, as our method consistently achieved better results.
Prefix tuning showed further limitations, which modifies only domain-specific embeddings, suffered from catastrophic forgetting, where new prefixes interfered with prior knowledge.

In summary, our method demonstrated the best balance of parameter efficiency and task adaptation, surpassing LoRA, block expansion, prefix tuning and full finetuning in maintaining high accuracy, forward and backward transfer.

\subsubsection{Trade-off between model size and performance}

\begin{figure}[h]
    \centering
    \begin{subfigure}[b]{0.46\textwidth}
        \centering
        \includegraphics[width=\textwidth]{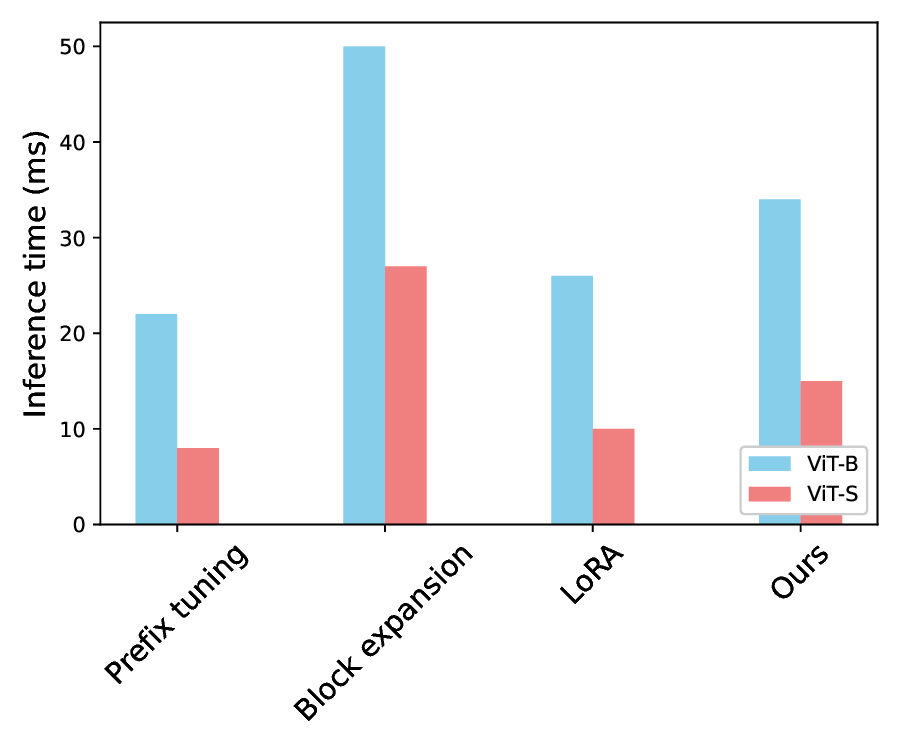} 
        \caption{Inference time comparison}
        
        \label{fig:inference-time}
    \end{subfigure}
    \hspace{0.00\textwidth}   
    \begin{subfigure}[b]{0.46\textwidth}
        \centering
        \includegraphics[width=\textwidth]{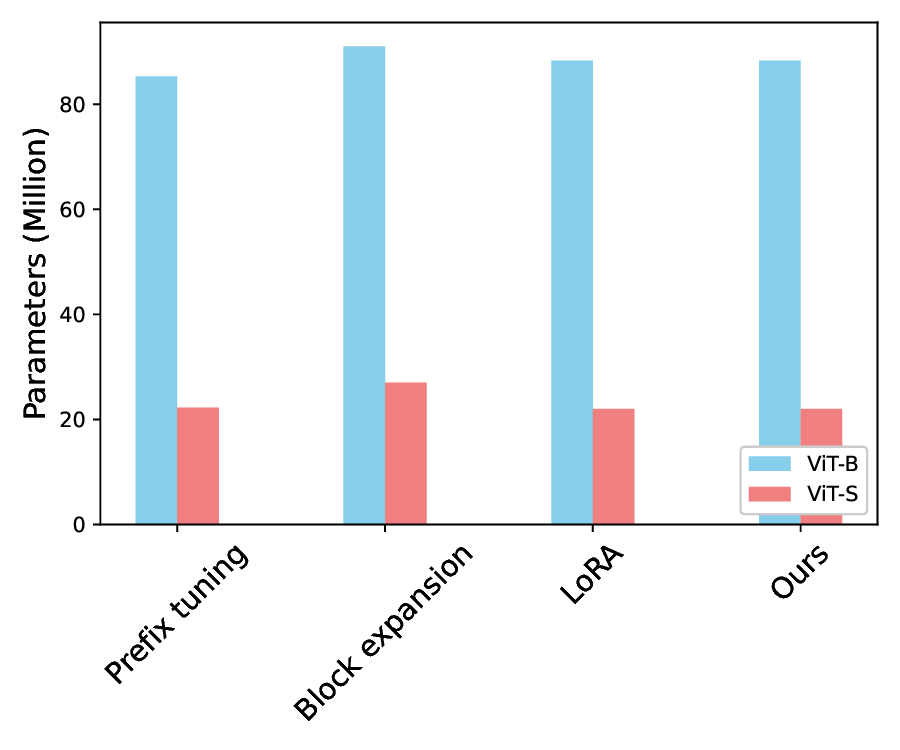} 
        \caption{Model size comparison}
        \label{fig:parameters}
    \end{subfigure}
    \caption{Comparison of efficiency for PEFTs methods.}
    \label{tab:big-small-comparison}
\end{figure}

As shown in Figure~\ref{tab:big-small-comparison}, the ViT-B model with 85 million parameters, offers higher accuracy but requires longer inference time than the ViT-S model, which has only 22 million parameters. This trade-off highlights the cost of higher performance, where larger models achieve superior accuracy but at a greater computational cost. In resource-limited scenarios, the ViT-S model often becomes the practical choice, balancing reasonable accuracy with efficiency.
Our proposed architecture further explores this by comparing LoRA with single-output and our method with multiple-output. Multiple-output configurations, which generate intermediate predictions, slightly increase inference time but it offers more performance and flexibility for continual learning. 

To further investigate this trade-off (efficiency vs. performance), we compare our approach with prefix tuning and block expansion techniques. Prefix tuning introduces only a small set of task-specific parameters to the input, is highly efficient, adding minimal parameters and maintaining faster inference times compared to other approaches. However, its performance is $20\%$ lower than our approach. Block expansion, in contrast, significantly increases the parameter count and inference time by adding a new block for each task, making it less suitable for real-time scenarios.

\section{Conclusion}
\label{section:conclusion}

In this work, we present a novel approach to continual learning using a dynamic architecture that incorporates LoRA to maintain parameter efficiency within ViTs. Our results demonstrate the effectiveness of domain-specific output heads and feature gating, respectively, in minimizing catastrophic forgetting and enabling the model to retain previously learned knowledge while integrating new domains. Through experiments, we highlight the significant impact of dataset order to preserves the accuracy, showing that training on fine-grained domains after generic ones enhances model performance and adaptability.

Future work should explore advanced strategies for incorporate larger-scale datasets, and address the challenges and drawbacks of adding new output heads for each dataset. Additionally, a key challenge at test time is determining which output head to use for a new domain. Furthermore, expanding the method to support continual self-supervised learning with domain-specific distributions, such as medical and geospatial images, could further enhance its applicability.

\bigskip

\bibliography{article}
\end{document}